\documentclass{article}
\usepackage{spconf,amsmath,graphicx}
\ninept

\usepackage{graphicx}
\usepackage{amssymb} 
\usepackage{color}

\usepackage{graphicx}
\usepackage[shortlabels]{enumitem}
\usepackage{tikz}
\usepackage{comment}
\usepackage{amsmath,amssymb} 
\usepackage{color}

\usepackage{subcaption}

\usepackage{booktabs}
\usepackage{multirow}
\usepackage{diagbox}
\usepackage{caption}
\usepackage{graphicx}
\usepackage{amsmath,amssymb}

\usepackage{pifont}

\usepackage{float}

\usepackage{caption}

\usepackage{subcaption}

\usepackage{booktabs}
\usepackage{amssymb,amsmath,nccmath}
\usepackage{cclicenses}
\usepackage{makecell}
\usepackage{lscape,array}
\newcolumntype{C}[1]{>{\centering\arraybackslash}p{#1}} 
\usepackage[thin, , thinc]{esdiff}
\usepackage{framed}

\usepackage{lmodern}
\usepackage{url}

\usepackage{gensymb}
\newcommand{\cmark}{\ding{51}}
\newcommand{\xmark}{\ding{55}}

\usepackage{graphicx}
\usepackage{amsmath}
\usepackage{amssymb}
\usepackage{booktabs}

\usepackage{gensymb}

\title{mmSense: Detecting Concealed Weapons with a Miniature Radar Sensor}
\name{\begin{tabular}{@{}c@{}}
Kevin Mitchell $^{\dagger}$ $^{\mathsection}$ \sthanks{We acknowledge funding from the QuantIC Project funded by the EPSRC Quantum Technology Programme (grant EP/MO1326X/1), and Google.} \qquad  Khaled Kassem $^{\dagger}$ $^{\mathsection}$ \qquad  Chaitanya Kaul $^{\ddagger}$ $^{\mathsection}$ \qquad \\ Valentin Kapitany $^{\dagger}$ \qquad  Philip Binner $^{\dagger}$ \qquad  Andrew Ramsay $^{\ddagger}$ \qquad \\ Roderick Murray-Smith $^{\ddagger}$ \qquad  Daniele Faccio $^{\dagger}$ \end{tabular}}

\address{$^{\dagger}$School of Physics and Astronomy, University of Glasgow, UK, G12 8SU \\
         $^{\ddagger}$ School of Computing Science, University of Glasgow, UK, G12 8RZ}

\begin{document}
%
{\maketitle}
\def\thefootnote{$\mathsection$}\footnotetext{Equal Contribution}
\begin{abstract}
For widespread adoption, public security and surveillance systems must be accurate, portable, compact, and real-time, without impeding the privacy of the individuals being observed. Current systems broadly fall into two categories -- image-based which are accurate, but lack privacy, and RF signal-based, which preserve privacy but lack portability, compactness and accuracy. Our paper proposes {\it mmSense}, an end-to-end portable miniaturised real-time  system that can accurately detect the presence of concealed metallic objects on persons in a discrete, privacy-preserving modality. mmSense features millimeter wave radar technology, provided by Google's Soli sensor for its data acquisition, and TransDope, our real-time neural network, capable of processing a single radar data frame in 19 ms. mmSense achieves high recognition rates on a diverse set of challenging scenes while running on standard laptop hardware, demonstrating a significant advancement towards creating portable, cost-effective real-time radar based surveillance systems. 
\end{abstract}

\section{Introduction}
\label{sec:intro}
\begin{figure}[t]
\begin{center}
\includegraphics[width=1.0\linewidth]{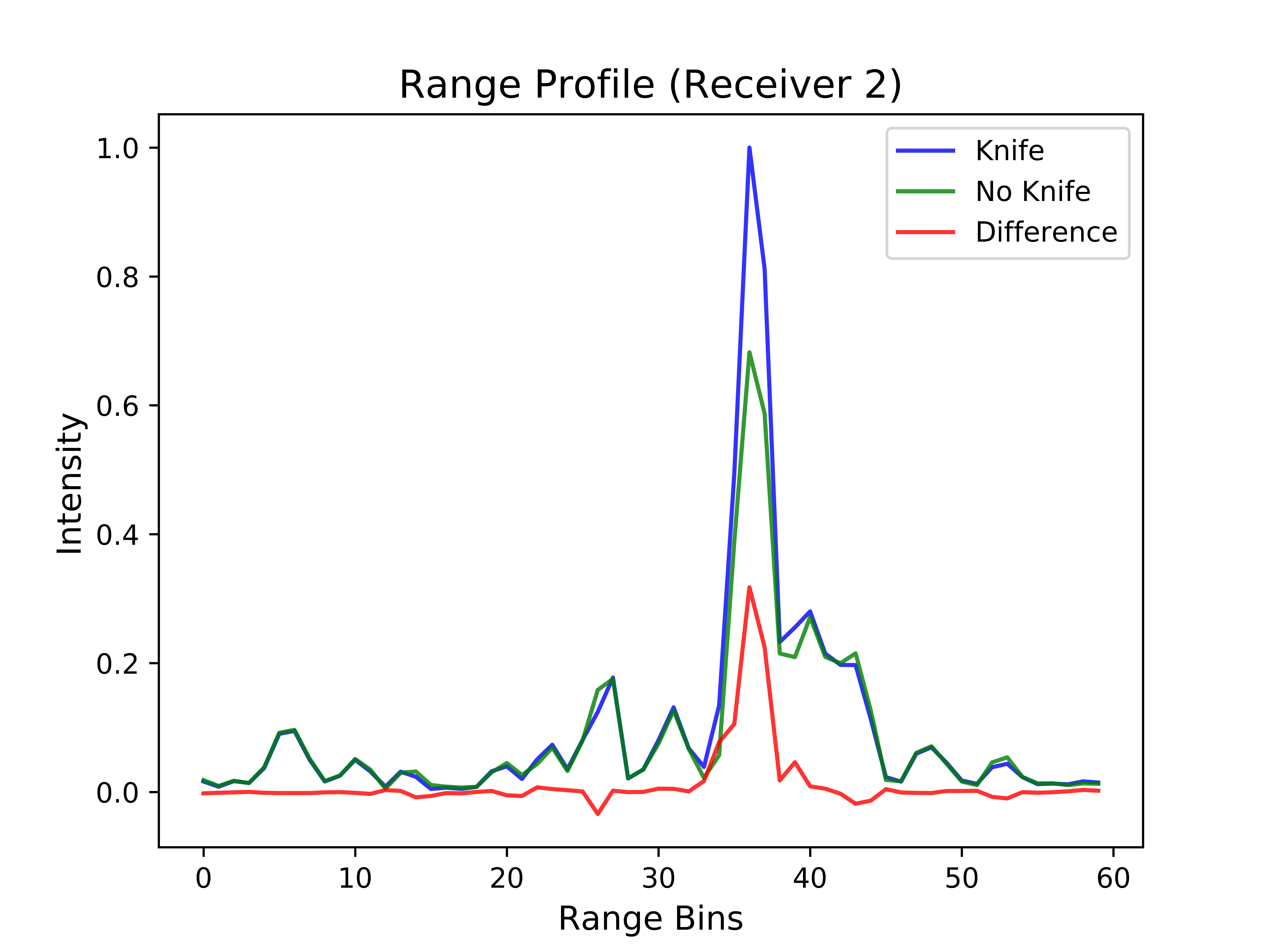}
\end{center}
\vspace{-0.5cm}
\caption{ The RF waves from a Google Soli reflect from a person standing $1.5$m away from the radar with and without a knife. To mitigate the variations in specular reflection, the subject rotated through 90$^{\circ}$ and an average over 60s was used in each scene. The reflected signal received via receiver 2 (of 3) of the Soli is converted into a range profile by computing an FFT over it, and plotted here. The difference between the two range profiles shows the potential of using mmWave radars like the soli, for detecting metallic objects.}
\label{proofofconcept}
\end{figure}

Radar solutions developed on Frequency Modulated Continuous Wave (FMCW) technology have shown promising success through their ability to serve as a capable and versatile basis for computational sensing and short range wireless communication systems \cite{solifabric}. Such radars, operating at millimeter wave (mmWave) frequency, can be used for robust gesture generation and recognition \cite{soliinteracting, radarnet}, and even measure distances with \textit{mm} accuracy \cite{soidistancesense}. Furthermore, mmWave radars have the potential to serve as a basis for concealed metallic object detection (e.g. knives, guns etc) which presents a novel and most importantly, privacy-preserving manner of real-time surveillance. The principles of mmWave metal detection rely on the underlying physics of RF waves- radio frequency (RF) waves that fall in the $30$--$300$ GHz range between microwaves and terahertz waves. This frequency band corresponds to wavelengths of $1$--$10$ mm. Within various forms of spectral imaging (e.g. IR, UV), one chooses the waveband that interacts with the object/scene to be imaged, whilst ignoring any obfuscating features. The same is true when detecting metals concealed on humans, where the mmWave waveband is appropriate because the waves from a mmWave RF source pass through the thin layers of clothes, and are reflected highly by the body, plus any hidden objects between the two \cite{Bjarnason2004}. Figure \ref{proofofconcept} depicts this concept empirically. We believe there is a niche in the security field for a portable technology that can screen for illegal metallic weapons, whilst still allowing people to maintain their freedom and privacy in public without security bottlenecks and conventional image-capturing cameras. 

This work is not the first to propose mmWave sensing for metal detection. Fusion of mmWave radar and vision data has already helped create accurate and effective object detection systems for autonomous driving applications \cite{mmwaveautonomous}. The comparison in performance of Convolutional and Recurrent Neural Networks (CNNs and RNNs) for metal detection using magnetic impedance sensors has been extensively evaluated in \cite{metaldetmi}. Their system is compact, however they need to scan the complete scene, and restrict themselves to large metal sheets that are visible in the Line-of-Sight (LOS) of their sensor. \cite{mmforsight} use mmWave radar sensors to alert robots to the presence of humans. Of most relevance to our study are \cite{metaldetstudy, aimetdetbeforeus}- the former study, \cite{metaldetstudy} provides a comprehensive guide on the metal detection capabilities of a $77-81$GHz radar chip but do so by comparing intensities of the reflected signal with the intensities of their own model of a body without the presence of the metallic object. Their work does not look at concealed metallic objects but ones that are already visible. \cite{aimetdetbeforeus} created an AI powered metal detection system capable of working in real-time. The system, however, processes their data differently, is prohibitively expensive, and is considerably bulkier than ours. One fundamental advantage our system has over all existing mmWave systems proposed for similar applications is the use of a radar sensor that has the widest Field of View (FOV), smallest form factor and least power consumption amongst its competitors. The Soli is capable of illuminating its surroundings with a $150^o$ FOV radar pulse. The lower power consumption of the Soli is due to the fact that it transmits 16 chirps at a pulse repetition frequency of $2000$ Hz in a single burst (each burst is transmitted at $25$Hz), and then stops transmitting until the next burst to save power \cite{radarnet}. This saves considerable power compared to mmWave radars used in existing works that continuously transmit chirps. Compared to current radar based surveillance systems, our technology does not need to sweep a scene to work, but provides inference on-the-fly by illuminating the scene with RF waves and processing the received signal.

\begin{figure*}[ht]
\begin{center}
\includegraphics[width=1.0\linewidth]{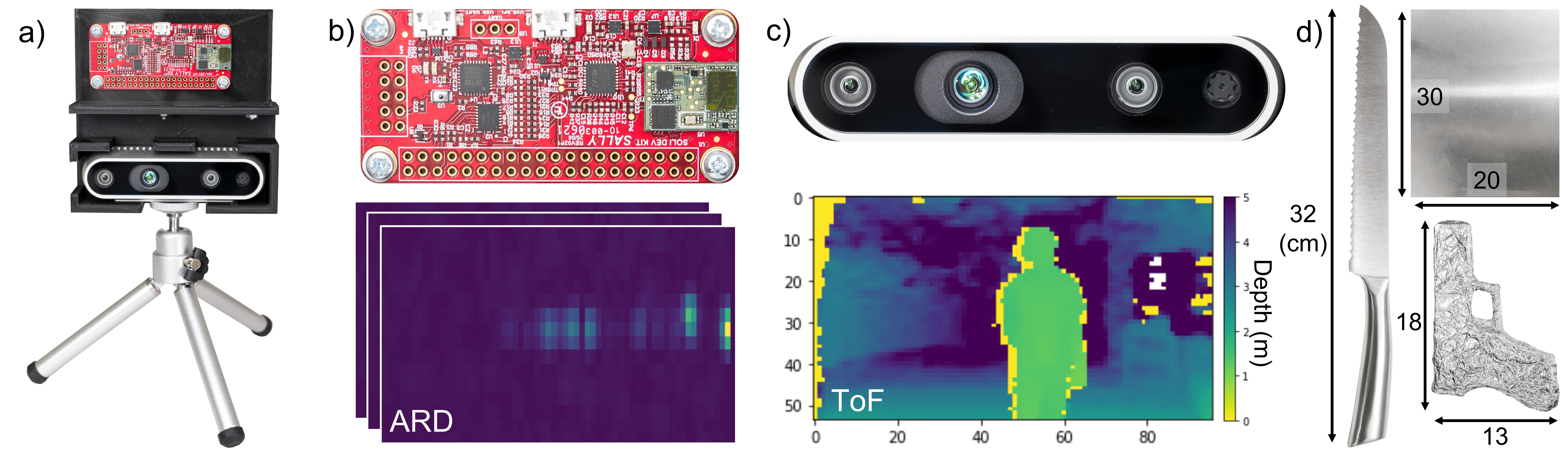}
\end{center}
\vspace{-0.5cm}
\caption{The \texttt{mmSense} set up is shown in (a). The individual components of the set up, and exemplar visualizations of the imaged scenes are shown for the Soli, and the Intel RealSense D435 in (b) and (c). The knife, gun, and metal sheet, used for the experiments are shown in (d).}
\label{setup}
\vspace{-0.3cm}
\end{figure*}

Our work is intended to disrupt the trend of specialised surveillance and imaging systems which are becoming increasingly expensive to install and operate, by using an inexpensive, compact device capable of being mounted in various locations throughout open spaces, which can function in real time. To this end, we present the use of a commercial mmWave radar transceiver to detect the presence of concealed objects on people in real time, in a privacy preserving manner. We focus on high frequency ($60$GHz), short range (up to $3$m) sensing using Google's Soli sensor, primarily due its miniature form factor, low power consumption, portability, and novel sensing characteristics. The Soli is designed for Human-Computer Interaction applications, and has shown success in 'macro' radar-based computational imaging tasks (detecting motion, gestures etc). Its application to detecting objects within the movement is unexplored and challenging. The Soli captures a superposition of reflected energy from different parts of a scene using high frequency RF waves: this results in poor lateral spatial resolution, while detecting even the smallest amount of motion. This makes metal detection challenging when there is plenty of movement in the scene i.e., in all practical real world scenarios. To mitigate this challenge, we propose a novel, real-time Vision Transformer model that can exploit semantic sequential relations in the preprocessed radar data and recognize the presence of a concealed metallic object on a person in a scene while ignoring objects such as wallets, keys, belts and mobile phones. 

The following are our main contributions: $(1)$ We present \texttt{mmSense} - a novel, end-to-end framework capable of detecting concealed metallic objects on people in a scene using only raw radar data without any human intervention. $(2)$ \texttt{mmSense} is real-time, and can potentially use off-the-shelf commercially available hardware making it easy to replicate and deploy. $(3)$ We open source \textit{mmSense} including all our datasets and models with the hopes of facilitating further research in this novel field of Artificial Intelligence powered concealed metal detection with mmWave RF signals.

\section{mmSense}
\label{sec:mmsense}

\begin{figure*}[ht]
\begin{center}
\includegraphics[width=1.0\linewidth, trim={0.5cm 13.5cm 0cm 8.2cm},clip]{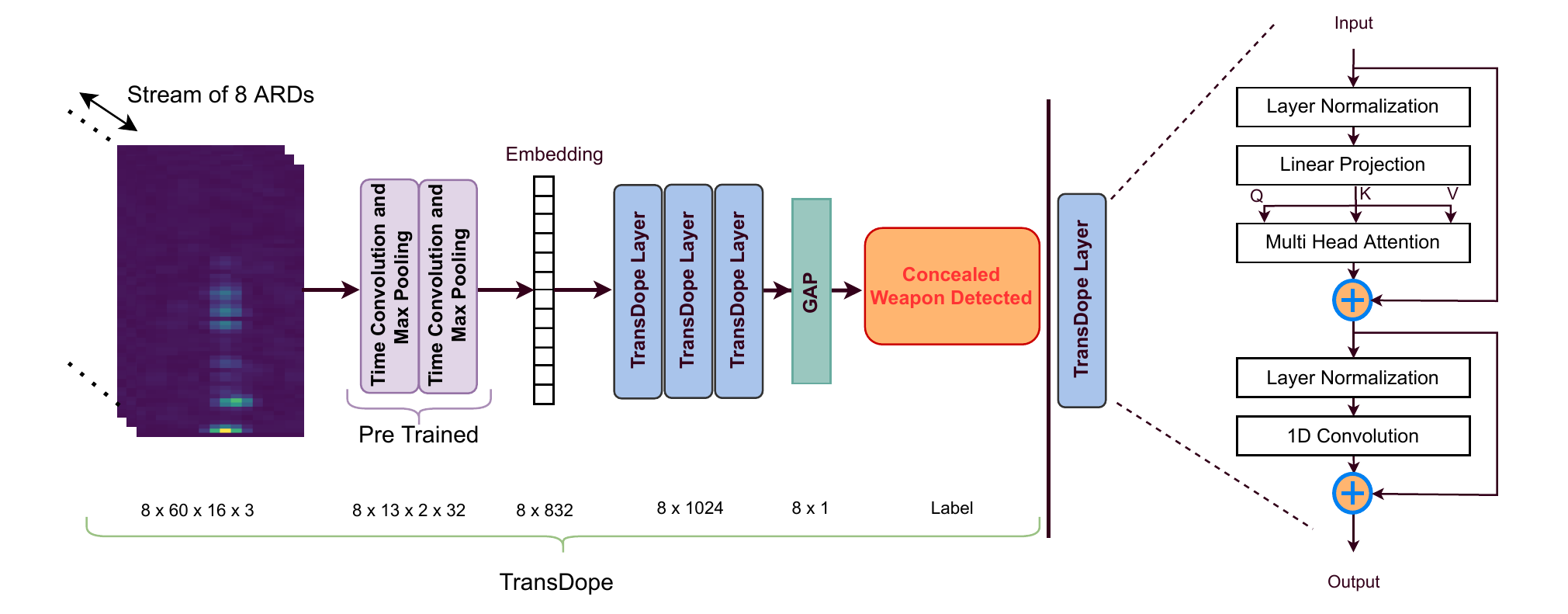}
\end{center}
\vspace{-0.5cm}
\caption{TransDope (shown here) processes the radar data stream while preserving time-dependent information throughout.}
\label{transdope}
\vspace{-0.3cm}
\end{figure*}

Our mmSense pipeline comprises of three components -- a Google Soli radar for data acquisition, an Intel RealSense D435 Time-of-Flight (TOF) camera for visualizing results, and an API capable of acquiring and processing radar data streams in real-time. A single burst of the Soli's transmitted signal is received across 3 antennas. For each RF illumination of the scene by the radar, it receives back a signal $I \in \mathbb{R}^{P \times C}$ where $I$ is the imaged scene as perceived by the radar and $P$ is the number of chirps received by the radar across its $C (=3)$ antennas. We operate the Soli at a frequency of $60$GHz with $1$GHz, the maximum permitted bandwidth $BW$. This gives us a range resolution $R_r = \frac{c}{2BW} = 15$cm, where $c$ refers to the speed of light. This is the minimum distance that the radar can separate in its LOS between two distinct points. The Soli transmits and receives a series of chirps that are grouped into bursts- we define the number of chirps for our system to be $16$, and collect bursts at $25$Hz, giving us 25 bursts of Soli data in one second. In this configuration, we can detect up to a maximum range of $9.6$m. 

The Soli hardware has a corresponding API provided by Google and implemented in C++. We built a C++ application around this API which allows us to interface with the Soli radar in real-time (e.g. selecting a range profile) and receiving the bursts generated from the radar. Our application supports streaming the Soli bursts directly to a Python script using the ZeroMQ$^1$ messaging framework. The bursts are relayed immediately upon being received by the device with no additional buffering, and are ultimately parsed by a Python module which extracts both the parameters associated with the burst (e.g. a hardware-supplied timestamp) and the raw radar data.
\def\thefootnote{1}\footnotetext{\url{https://github.com/zeromq/cppzmq}}

After parsing the raw chirps and their timestamps, we create a Range Doppler \cite{soliubiquitou} transformation of the signal. This is done via a series of Fast Fourier Transforms (FFT) applied to the data. First, we calculate the complex value range profile (RP) of the radar signal. This is done via an FFT of the radar chirps received by the 3 antennas. As the Soli's signal is a superposition of reflections from a scene, the RP data can be interpreted as how well the separate contributions of the RF scatters in the scene are resolved. This gives us an estimate of the geometry of the scene. A Complex Range Doppler (CRD) plot is then calculated as an FFT over each channel of the radar's complex value range profile. Here, the range represents the distance of the object in a scene from the Soli, and the Doppler corresponds to the radial velocity of the object towards the Soli. We use the magnitude of the CRD for our experiments, which is obtained in the following way, $\texttt{ARD}(r,d) = |\texttt{CRD}(r,d)|$, where $\texttt{ARD}$ refers to the Absolute Range Doppler, $r$ and $d$ are the range and doppler bins, and $|\cdot|$ is the absolute value.

The ARD plots are processed using our novel Deep Neural Network, TransDope (Doppler Transformer, figure \ref{transdope}), capable of accurate real time classification of the input data stream. The input is a sequence of 8 ARD frames. TransDope contains two Time Convolution layers pretraind on a large collected dataset of ARD plots, an embedding layer to create a transformer compatible embedding of the convolution features, and transformer layers to learn long range time-dependent semantic features in the data. We first collect a large dataset of ARDs from various scenes with and without concealed metallic objects on actors. We then train a model with two Time Convolution and Max Pooling layers, the output of which is flattened and fed to a classification layer. 

Following training, we discard the output layer, and use the two time convolution layers with the pre-trained weights as an initialization. Unlike standard convolutions that apply a weight kernel across all 8 ARDs concurrently, we apply convolutions sequentially to the 8 ARD frames to extract time-dependent features from them, and hence call them time convolutions. We then reshape the output of the last Max Pooling layer to create an embedding of the features. We also add positional encoding to each of the 8 ARD frames to preserve their sequence. Following this, we pass the embedding through 3 TransDope layers that extract semantic dependencies within the ARD's feature representation. These layers are the same as ViTs \cite{vit} encoder layers with the exception of having a convolutional layer following the multi head attention layer, instead of the dense layers, to reduce parameter size. We use global average pooling to reduce the transformer layer's features to a vector, which are then passed into the output layer. Our Time Convolution layers have $32$ filters and a kernel size of $3\times3$. Our transformer layer has an embedding size of $128$ and uses $2$ attention heads. TransDope contains $0.8$ million parameters, and can process a single ARD frame in $19$ milliseconds on an Intel i$9$ $8$-core CPU. We train our model in TensorFlow 2 for 50 epochs, with a batch size of $8$, and a learning rate of $1e-2$ which inversely decays after every 10 epochs. During inference, we feed $1$ batch of 8 ARD frames through the model to get a classification. 

\section{Experiments}
\label{sec:experiments}


\begin{table}[h]
  \centering
    \begin{tabular}{l|*{6}r}
    \toprule
    \diagbox{P}{S} & A & B & C & D & E & F\\
    \midrule
    Object & M & K & K & G & G & G\\
    \hline
    People & 1 & 5 & 1 & 2 & 1 & 1\\
    \hline
    Dist. (m) & 2.0 & 2.0 & 2.0 & 2.85 & 1.5 & 2.0 \\
    \hline
    Closed & \xmark & \xmark & \xmark & \xmark & \cmark & \cmark \\
    \hline
    Accessories & \cmark & \cmark & \xmark & \xmark & \xmark & \xmark\\
    \hline
    Acc. (\%) & 95.1 & 86.9 & 88.4 & 89.0 & 74.6 & 79.2\\
    \bottomrule
    \end{tabular}
  \caption{Distinct properties (P) of the scenes (S) we collected, along with the accuracy of TransDope to identify the object for that scene. Here, K, G and M refer to the Knife, Gun and Metal Sheet respectively, people is the number of individuals in the scene and distance (in metres) is the maximum distance from the Soli that the individuals in the scene walk up to. Closed refers to a scene where the walls are close to the radar, and the RF signal can bounce off adjacent walls. Accessories are everyday items belts, wallets, keys and phones that the people in the scene carry.}
  \label{tab:datasets}
  \vspace{-0.6cm}
\end{table}
\vspace{-0.3cm}
To test the accuracy and flexibility of our technique, we collected 6 different scenes with varying characteristics, as depicted in table \ref{tab:datasets}. Data was acquired in 4 instances for each scene: 2 with a metallic object hidden on a person, and 2 without. Each acquisition contains approximately 1500 frames of data. Before training our machine learning model, we collected roughly 15,000 frames of Soli data equally split into the two classes in various scenes, to pre-train the TransDope time convolutions. For each scene, we then trained TransDope, to predict a binary class for each ARD. 

\begin{figure}[h]
   \centering
  \sbox0{\includegraphics[width=3.2cm,height=3.2cm,keepaspectratio,trim={2.5cm 2.5cm 2.5cm 2.5cm},clip]{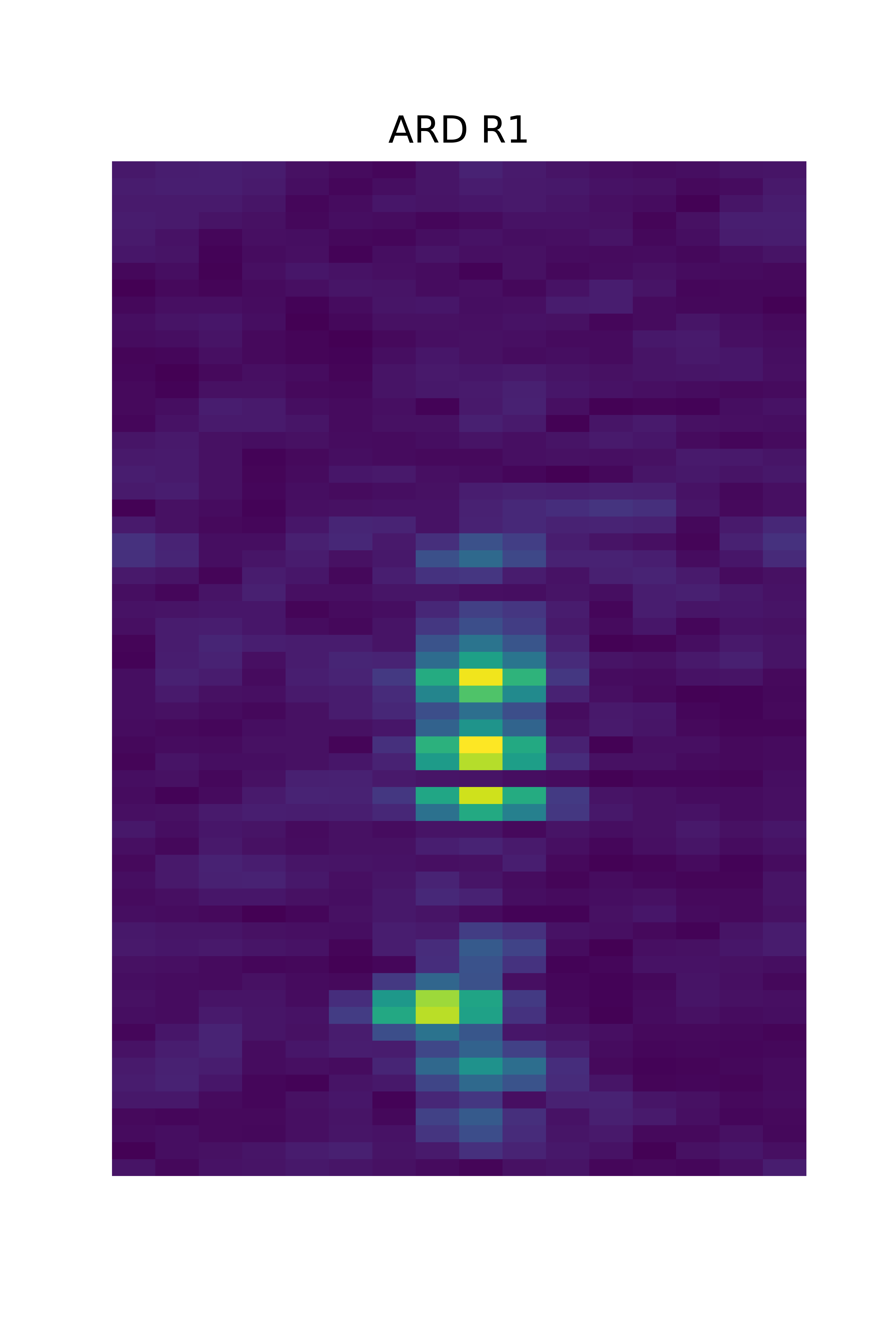}}%
  \sbox1{\includegraphics[width=3.2cm,height=3.2cm,keepaspectratio,trim={2.5cm 2.5cm 2.5cm 2.5cm},clip]{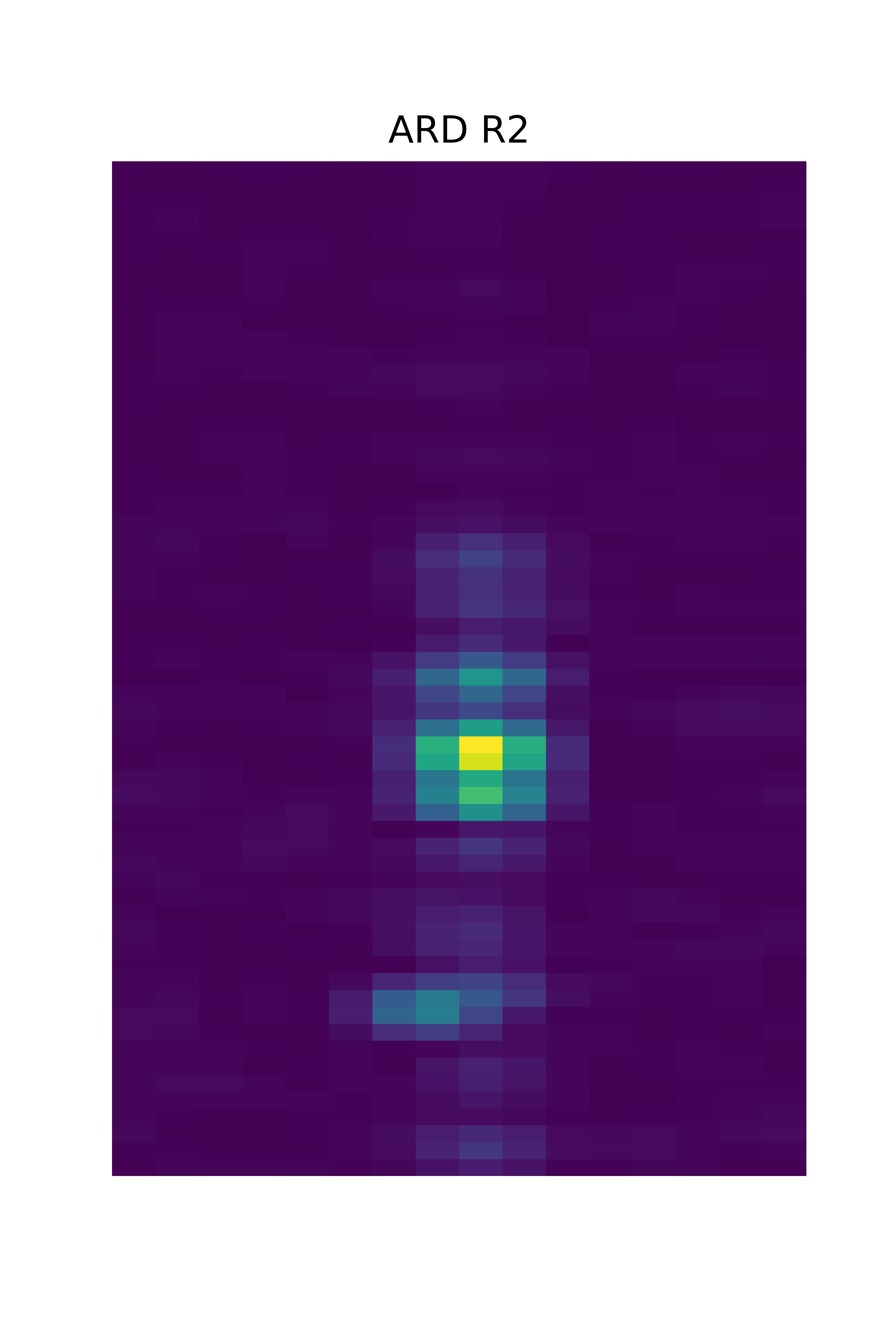}}%
  \sbox2{\includegraphics[width=3.2cm,height=3.2cm,keepaspectratio,trim={2.5cm 2.5cm 2.5cm 2.5cm},clip]{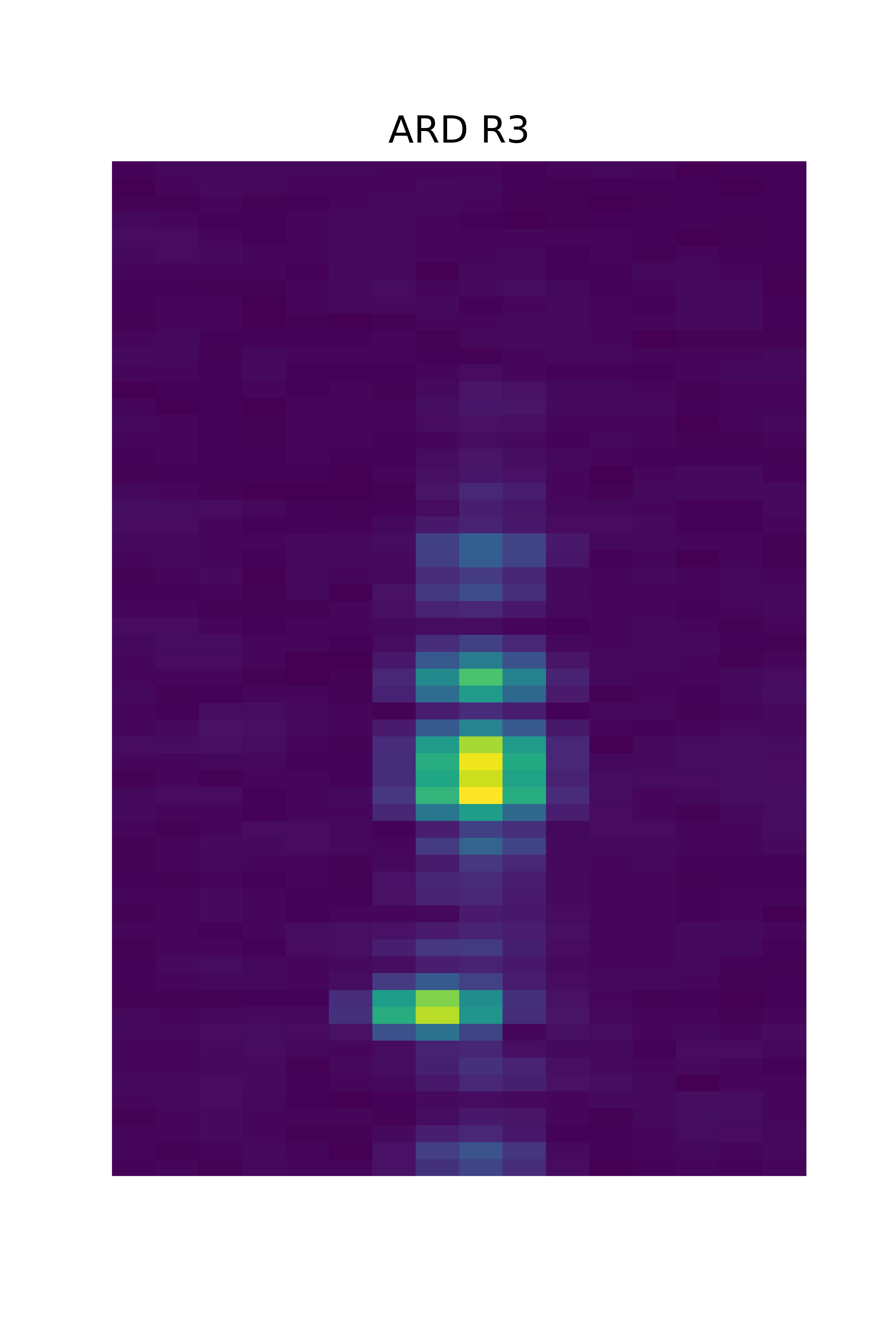}}%
  \sbox3{\includegraphics[width=3.7cm,height=3.2cm,,trim={1.5cm 1.5cm 1.5cm 1.5cm},clip]{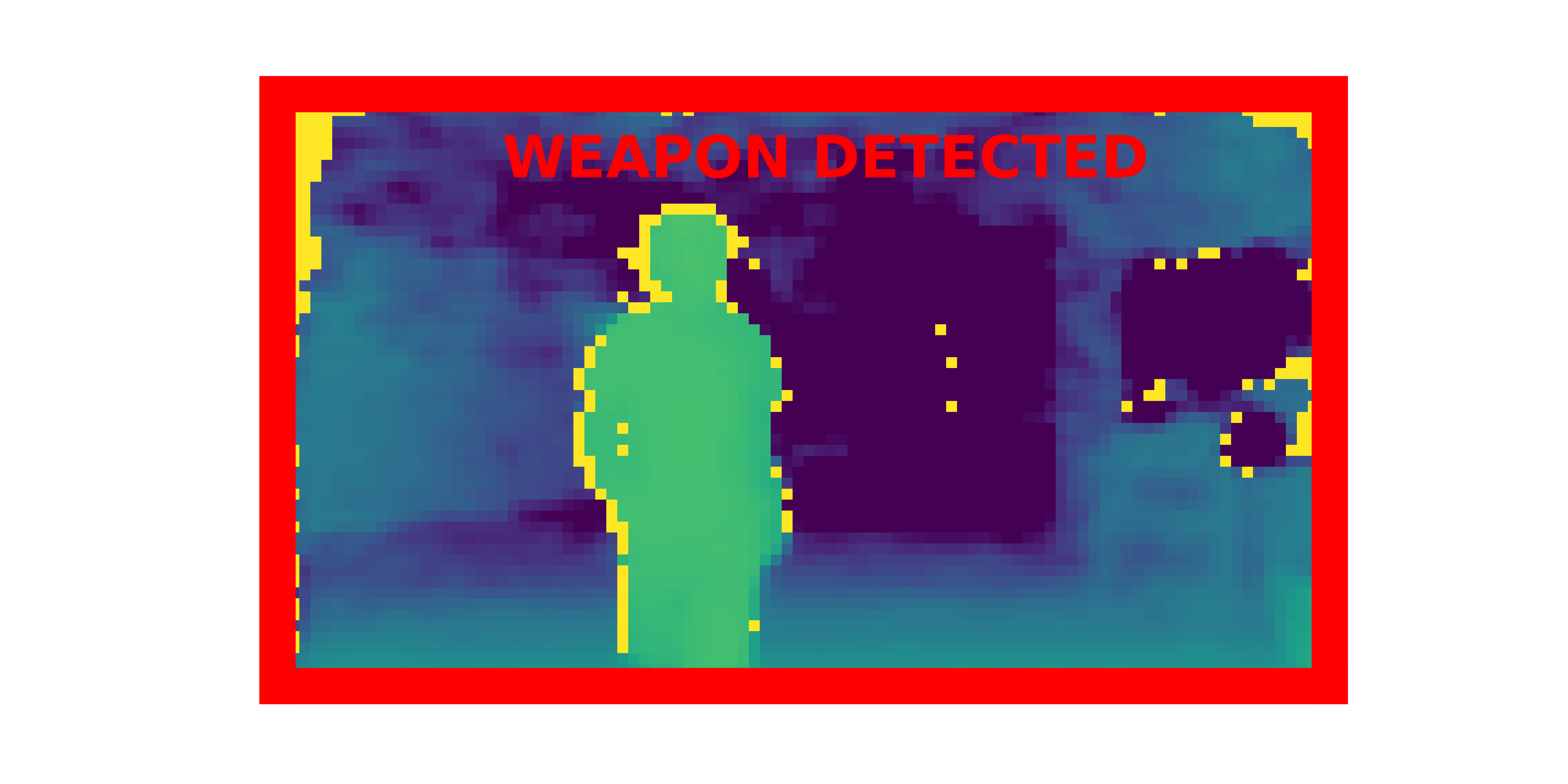}}%
  \begin{subfigure}{\wd0}\usebox0\end{subfigure}%
  \begin{subfigure}{\wd1}\usebox1\end{subfigure}%
  \begin{subfigure}{\wd2}\usebox2\end{subfigure}%
  \begin{subfigure}{\wd3}\usebox3\end{subfigure}\\
  \sbox0{\includegraphics[width=3.2cm,height=3.2cm,keepaspectratio,trim={2.5cm 2.5cm 2.5cm 2.5cm},clip]{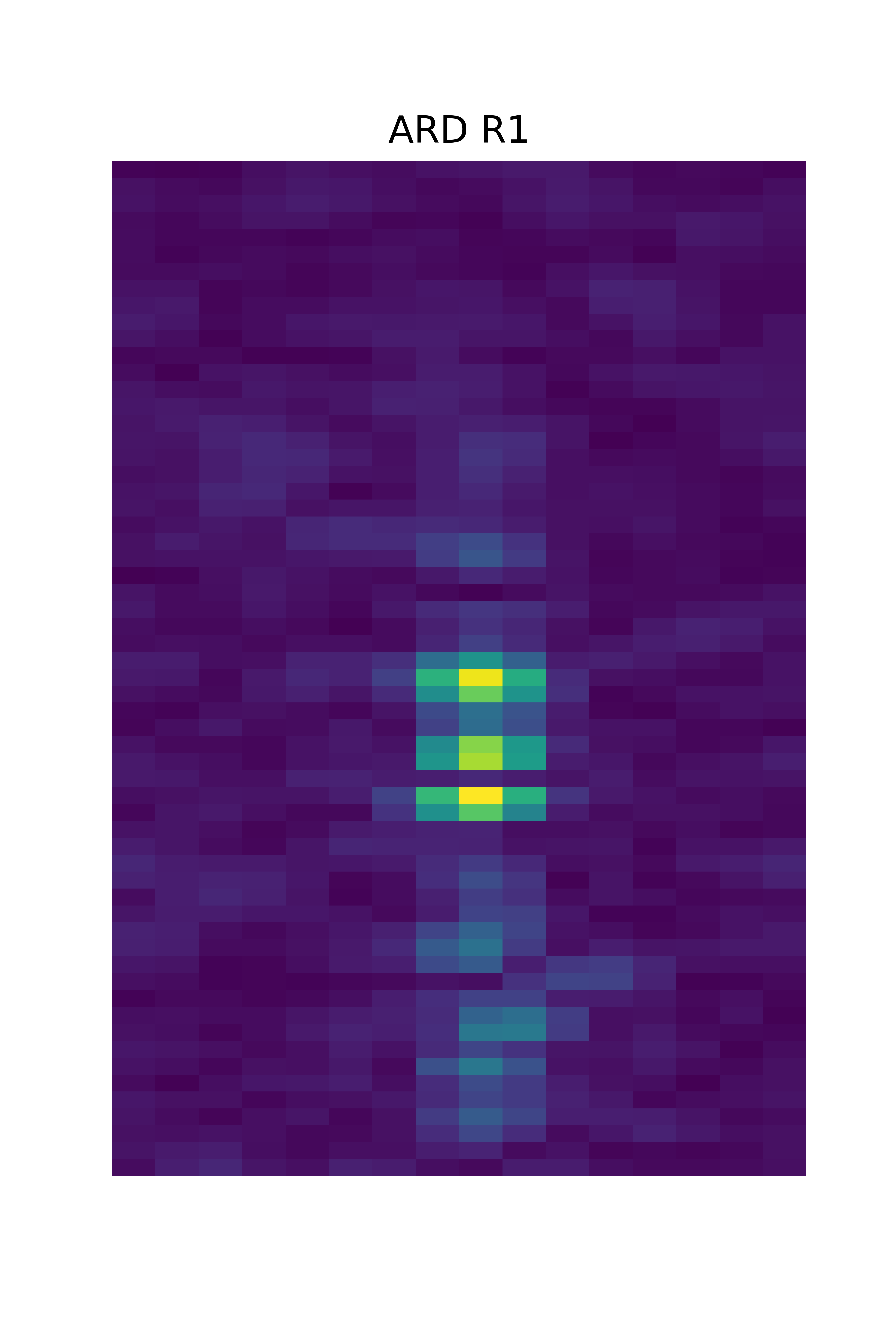}}%
  \sbox1{\includegraphics[width=3.2cm,height=3.2cm,keepaspectratio,trim={2.5cm 2.5cm 2.5cm 2.5cm},clip]{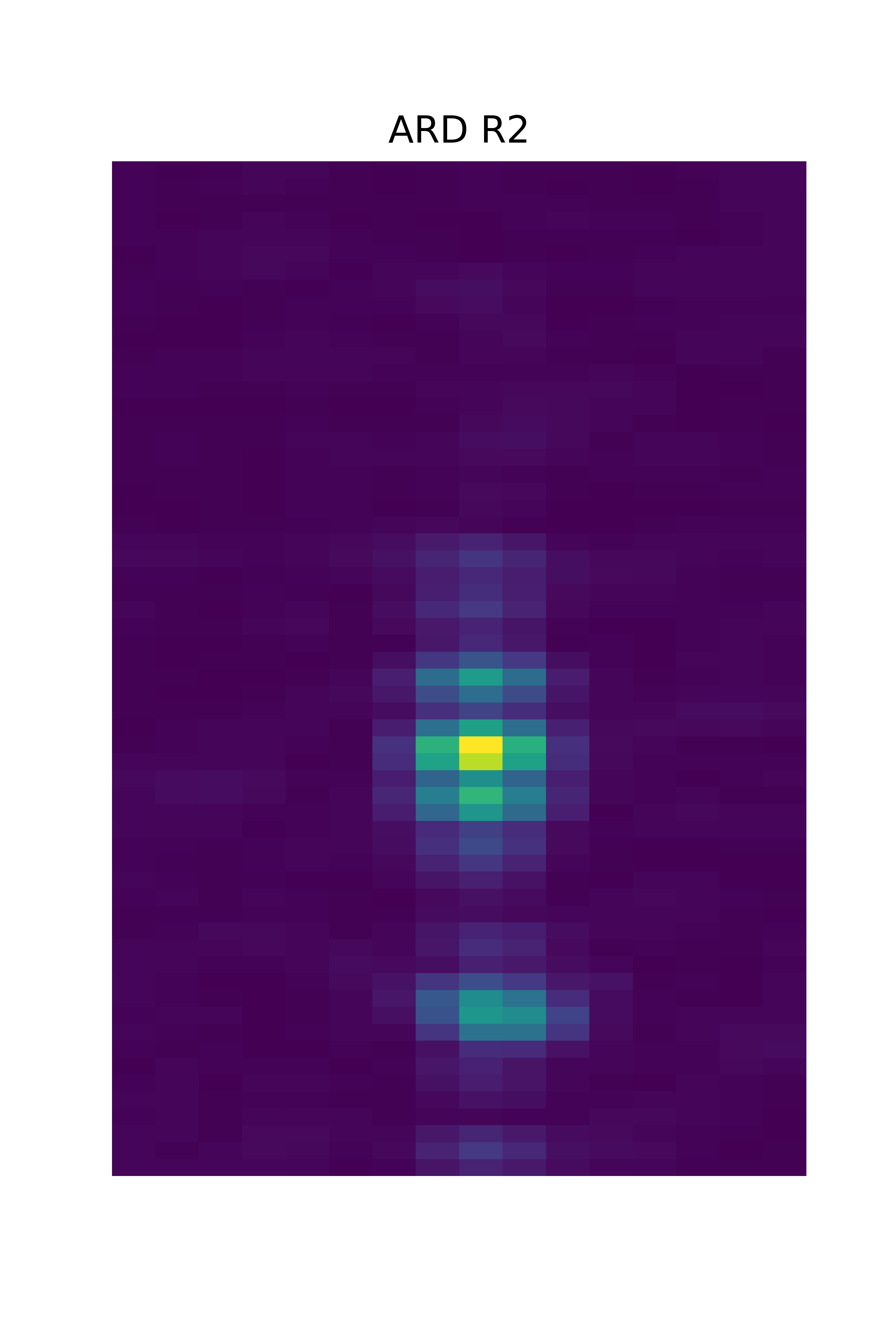}}%
  \sbox2{\includegraphics[width=3.2cm,height=3.2cm,keepaspectratio,trim={2.5cm 2.5cm 2.5cm 2.5cm},clip]{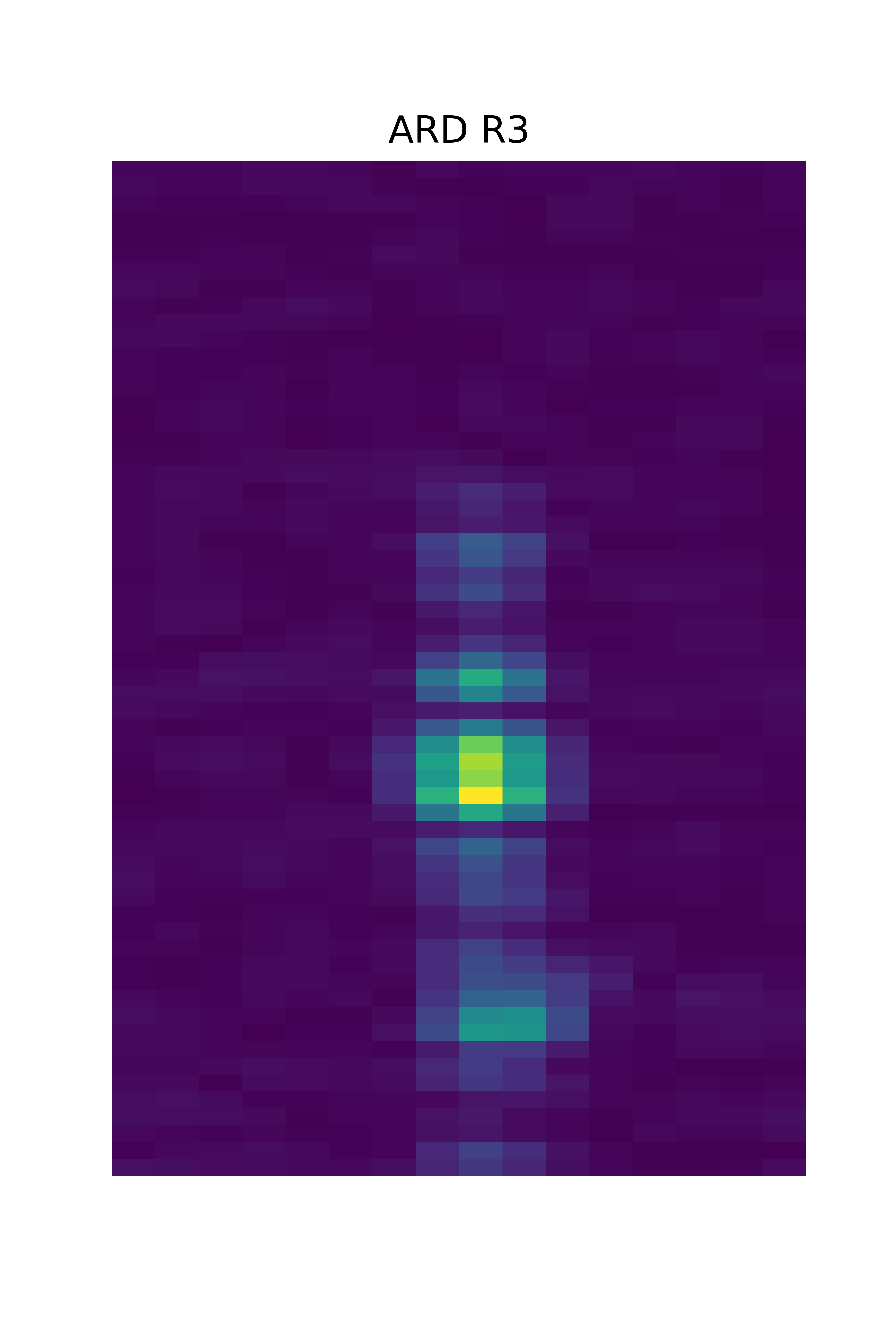}}%
  \sbox3{\includegraphics[width=3.7cm,height=3.2cm,trim={1.5cm 1.5cm 1.5cm 1.5cm},clip]{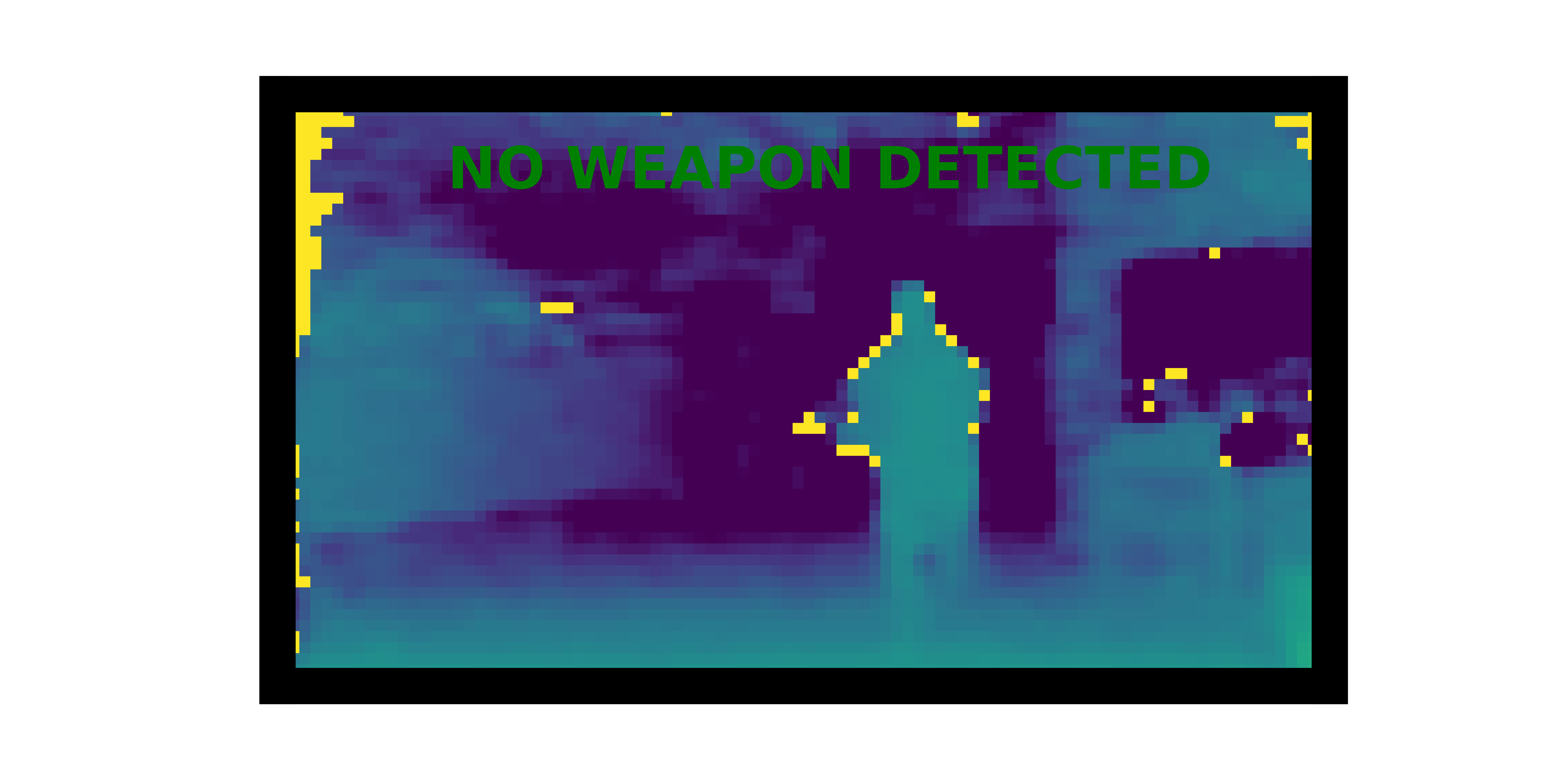}}%
  \begin{subfigure}{\wd0}\usebox0\end{subfigure}%
  \begin{subfigure}{\wd1}\usebox1\end{subfigure}%
  \begin{subfigure}{\wd2}\usebox2\end{subfigure}%
  \begin{subfigure}{\wd3}\usebox3\end{subfigure}\\
  \caption{Results for Scene C where the person has a knife hidden under a jacket. On the left are the ARDs for the three receiver antennas, and on the right is the output visualization. The red TOF output denotes the knife being detected, and the black is the misclassification due to the specular reflection from the knife not providing a strong signal.}
  \label{visualization}
\end{figure}

We carefully curated different scenes to portray real world situations where our system can be deployed. Scene A was an initial proof of concept where we used the metal sheet as the hidden object to verify the capabilities of our system. We were able to predict the presence of the sheet with $95.1\%$ accuracy. Scene B replicates a crowded expansive scenario, such as an airport terminal. Here we crowded the scene with 5 people walking up to 2m away in radius from the setup. Each person in the scene was carrying everyday objects such as phones, keys, wallets, and belts; only one of these individuals had a knife on their person. Even in such a challenging setting, our system detected the presence of the knife with up to $86.9\%$ accuracy. In Scenes C and D, we observed the effects of changing the hidden object from a knife to a gun. This is important as different objects have different characteristic specular reflections. As seen from our results, the performance of our system held when switching the metallic object from the knife to the gun. Scenes A to D were all open scenes, i.e. the data was not acquired with constricting walls. This results in no multipath RF signals received by the Soli receiving antennas. In Scenes E and F, we tested the effects of keeping our set up in a closed setting and noticed performance decreased. Our results are summarized in table \ref{tab:datasets} and visualized in figure \ref{visualization}.

\begin{figure}[H]
\begin{center}
\includegraphics[width=0.45\textwidth]{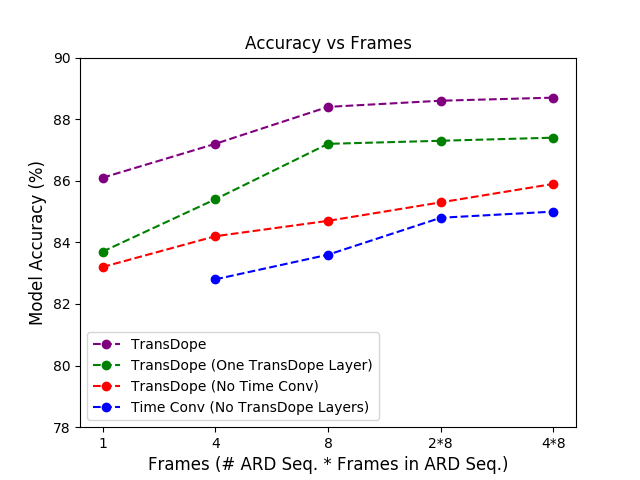}
\end{center}
\vspace{-0.4cm}
\caption{Effects of varying the number of frame sequences and model layers on performance on Scene C.}
\label{tab:ablations}
\vspace{-0.5cm}
\end{figure}

\textbf{Ablations.} Table \ref{tab:ablations} shows the effect of varying the amount of sequential information provided to TransDope, as well as varying the various blocks of TransDope. The experiments show that each individual component in our model contributes to performance boosts in terms of metal detection accuracy. We chose $8$ ARD frames per sequence as the input to TransDope due to it providing the best accuracy versus execution time. Having multiple sequences of ARDs does further boost performance, but it also doubles (for 2 ARD sequences of 8 ARD frames - $2*8$), and quadruples (for $4*8$) the execution time for only minor gains in performance. 

\vspace{-0.8cm}
\vspace{-0.5cm}
\section{Conclusions and Limitations}
\label{sec:experiments}
\vspace{-0.2cm}
Our paper proposed \textit{mmSense}, an AI assisted concealed metal detection system whose technology is centred on an mmWave transceiver. \textit{mmSense} can detect the presence of concealed weapons capable of mass harm even when scenes are crowded. It does so in real-time, and without compromising the privacy of the individuals in the scene. Our system however, has certain limitations in terms of performance degradation due to the effects of multipath RF waves at the receiver that occurs in close walled areas, as well as the inherent lack of spatial context from which commercial mmWave sensors suffer. We believe that investigating multi-modal sensor fusion of the radar and TOF data to add spatial awareness to the 'sensing' ability of \textit{mmSense} may help to alleviate these drawbacks, and would be a fitting next step to extend this technology.

\bibliographystyle{IEEEbib}
\bibliography{refs}

\end{document}